\def\year{2019}\relax
\documentclass[letterpaper]{article} 
\usepackage{aaai19}  
\usepackage{times}  
\usepackage{helvet}  
\usepackage{courier}  
\usepackage[hyphens]{url}  
\usepackage{graphicx}  
\usepackage{array}
\usepackage{amsmath}
\usepackage{amssymb}
\usepackage{latexsym}
\usepackage{textcomp}

\usepackage{color}
\usepackage{multicol}
\usepackage{multirow}
\usepackage{array}
\usepackage{arydshln}
            
\usepackage{helvet}
\usepackage{courier}

\usepackage{bm}
\usepackage{color}

\usepackage{subfig}
\usepackage{arydshln}

\usepackage{amsfonts}

\title{Context-Aware Self-Attention Networks}

\author{Baosong Yang$^1$~~~~~Jian Li$^2$~~~~~Derek F. Wong$^{1}$~~~~~Lidia S. Chao$^1$~~~~Xing Wang$^3$~~~~~Zhaopeng Tu$^3$\thanks{Zhaopeng Tu is the corresponding author. Work was done when Baosong Yang and Jian Li were interning at Tencent AI Lab.}\\
\bgroup

\begin{tabular}{ccc}
    \multicolumn{3}{c}{$^1$NLP$^2$CT Lab, Department of Computer and Information Science, University of Macau} \\
    \multicolumn{3}{c}{\tt nlp2ct.baosong@gmail.com, \{derekfw,lidiasc\}@umac.mo} \\
    $^2$The Chinese University of Hong Kong &   & $^3$Tencent AI Lab \\
    {\tt jianli@cse.cuhk.edu.hk}    &   \qquad \qquad \qquad   &   {\tt \{brightxwang,zptu\}@tencent.com} \\
\end{tabular}
\egroup
}

\begin{document}
\maketitle
\begin{abstract}
 Self-attention model  have shown its flexibility in parallel computation and the effectiveness on modeling both long- and short-term dependencies. However, it calculates the dependencies between representations without considering the contextual information, which have proven useful for modeling dependencies among neural representations in various natural language tasks. 
In this work, we focus on improving self-attention networks through capturing the richness of context.
To maintain the simplicity and flexibility of the self-attention networks, we propose to contextualize the transformations of the query and key layers, which are used to calculates the relevance between elements. 
Specifically, we leverage the internal representations that embed both global and deep contexts, thus avoid relying on external resources. 
Experimental results on WMT14 English$\Rightarrow$German and WMT17 Chinese$\Rightarrow$English translation tasks demonstrate the effectiveness and universality of the proposed methods. Furthermore, we conducted extensive analyses to quantity how the context vectors participate in the self-attention model.

\end{abstract}

\section{Introduction}

Self-attention networks (SANs) \cite{lin2017structured} have 
shown promising empirical results in various NLP tasks, such as machine translation~\cite{Vaswani:2017:NIPS}, nature language inference~\cite{Shen:2018:AAAI}, and acoustic modeling~\cite{sperber2018self}. 
One strong point of SANs is the strength of capturing long-range dependencies by explicitly attending to all the signals, which allows the model to build a direct relation with another long-distance representation. 

However, SANs treat the input sequence as a bag-of-word tokens and each token individually performs attention over the bag-of-word tokens. 
Consequently, the contextual information is not taken into account in the calculation of dependencies between elements.
Several researchers have shown that contextual information can enhance the ability of modeling dependencies among neural representations, especially for the attention models.
For example, \citeauthor{Tu:2017:TACL}~\shortcite{Tu:2017:TACL} and \citeauthor{zhang2017context}~\shortcite{zhang2017context} respectively enhanced the query and memory of a standard attention model~\cite{bahdanau2015neural} with internal contextual representations.
~\citeauthor{Wang:2017:EMNLP}~\shortcite{Wang:2017:EMNLP} and~\citeauthor{Voita:2018:ACL}~\shortcite{Voita:2018:ACL} enhanced the two components with external contextual representations that summarizes previous source sentences.

In this work, we propose to strengthen SANs through capturing the richness of context, and meanwhile maintain their simplicity and flexibility. 
To this end, we employ the internal representations as context vectors, thus avoid relying on external resources, e.g. the embeddings of previous sentences.
Specifically, we contextualize the transformations from the input layer to the query and key layers, which are used to calculate the relevance between elements. 
We exploit several strategies for the contextualization, including: 1) {\em global context} that represents the global information of a sequence; 2) {\em deep context} that embeds syntactic and semantic information summarized by multiple-layer representations; and 3) {\em deep-global context} that combines information of both the above two context vectors. 

Some researchers may doubt that, for a multi-layer self-attentive model (e.g. \textsc{Transformer} \cite{Vaswani:2017:NIPS}), each input state has summarized the global information from its lower layer through the weighted sum operation. Our study dispels the doubt by showing that such summarization does not fully captures the richness of contextual information.
We conducted experiments on two widely-used WMT14 English$\Rightarrow$German and WMT17 Chinese$\Rightarrow$English translation tasks.
The proposed approach consistently improves translation performance over the strong \textsc{Transformer} baseline, while only marginally decreases the speed.
Extensive analyses reveal that there exists separate requirement of contextual information for different representations, e.g.,  
the representation of the function words distinctly require more contexts than that of the content words.

\section{Background}
Recently, as a variant of attention model, self-attention networks~\cite{lin2017structured} have attracted a lot of interests due to their flexibility in parallel computation and modeling both long- and short-term dependencies. SANs calculate attention weights between each pair of tokens in a single sequence, thus can capture long-range dependency more directly than their RNN counterpart.

Formally, given an input layer ${\bf H}=\{h_1, \dots, h_n\}$, the hidden states in the output layer are constructed by attending to the states of input layer. Specifically, the input layer ${\bf H} \in \mathbb{R}^{n\times d}$ is first transformed into queries ${\bf Q} \in \mathbb{R}^{n\times d}$, keys ${\bf K} \in \mathbb{R}^{n\times d}$, and values ${\bf V} \in \mathbb{R}^{n\times d}$:
\begin{equation}
  \begin{bmatrix}
    \bf Q\\\bf K\\\bf V
  \end{bmatrix} = {\bf H}
  \begin{bmatrix}
    {\bf W}_{Q}\\{\bf W}_{K}\\{\bf W}_{V}
  \end{bmatrix},
\label{eq:linear}
\end{equation}
where $\{{\bf W}_{Q}, {\bf W}_{K}, {\bf W}_{V}\} \in \mathbb{R}^{d \times d}$ are trainable parameter matrices with $d$ being the dimensionality of input states. 
The output layer ${\bf O} \in \mathbb{R}^{n\times d}$ is constructed by
\begin{eqnarray}
   {\bf O} &=& \textsc{Att}({\bf Q}, {\bf K}) \  {\bf V} \label{eq:out},
\end{eqnarray}
where $\textsc{Att}(\cdot)$ is an attention model, which can be implemented as either additive attention~\cite{bahdanau2015neural} or dot-product attention~\cite{luong2015effective}.
In this work, we use the latter, which achieves similar performance with its additive counterpart while is much faster and more space-efficient in practice~\cite{Vaswani:2017:NIPS}:
\begin{eqnarray}
    \textsc{Att}(\bf Q, \bf K) &=& softmax(\frac{{\bf Q} {\bf K}^T}{\sqrt{d}}), \label{eq:sim}
\end{eqnarray}
where $\sqrt{d}$ is the scaling factor.

\subsection{Motivation}
\begin{figure*}[t]
\begin{center}
\subfloat[Conventional SANs]{\includegraphics[width=0.7\columnwidth]{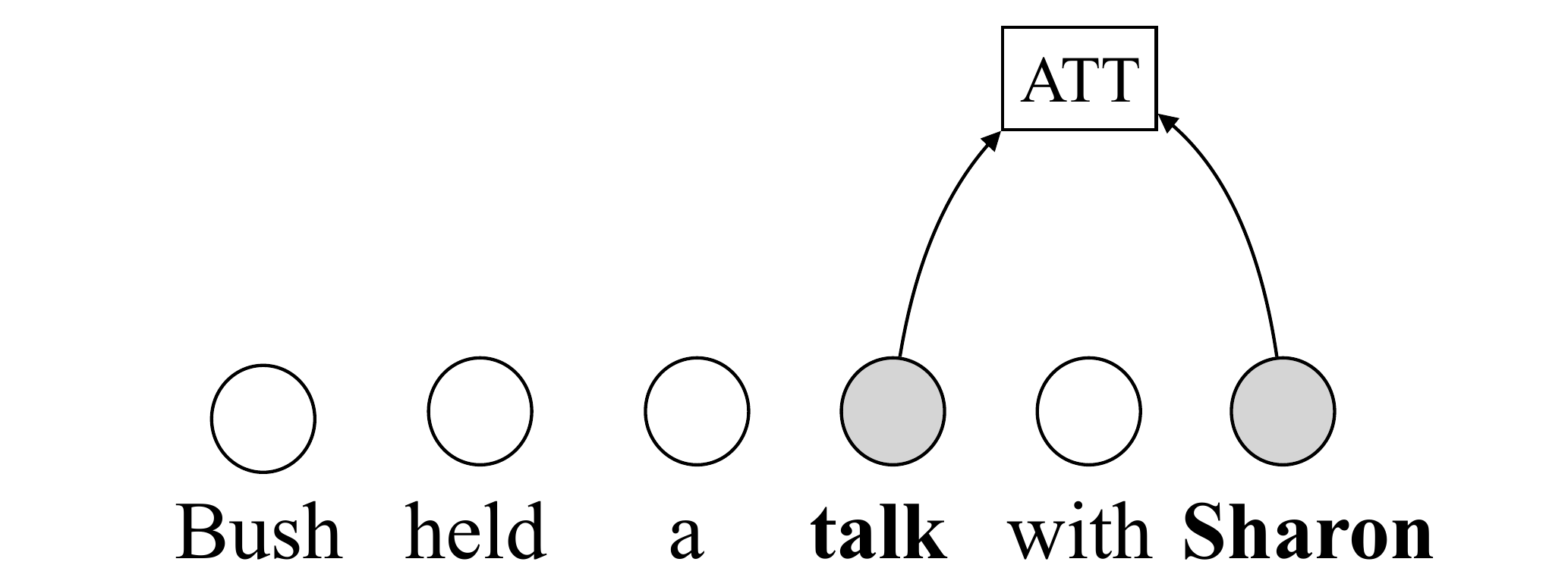}}\hspace{0.2\columnwidth}
\subfloat[Global Context]{\includegraphics[width=0.7\columnwidth]{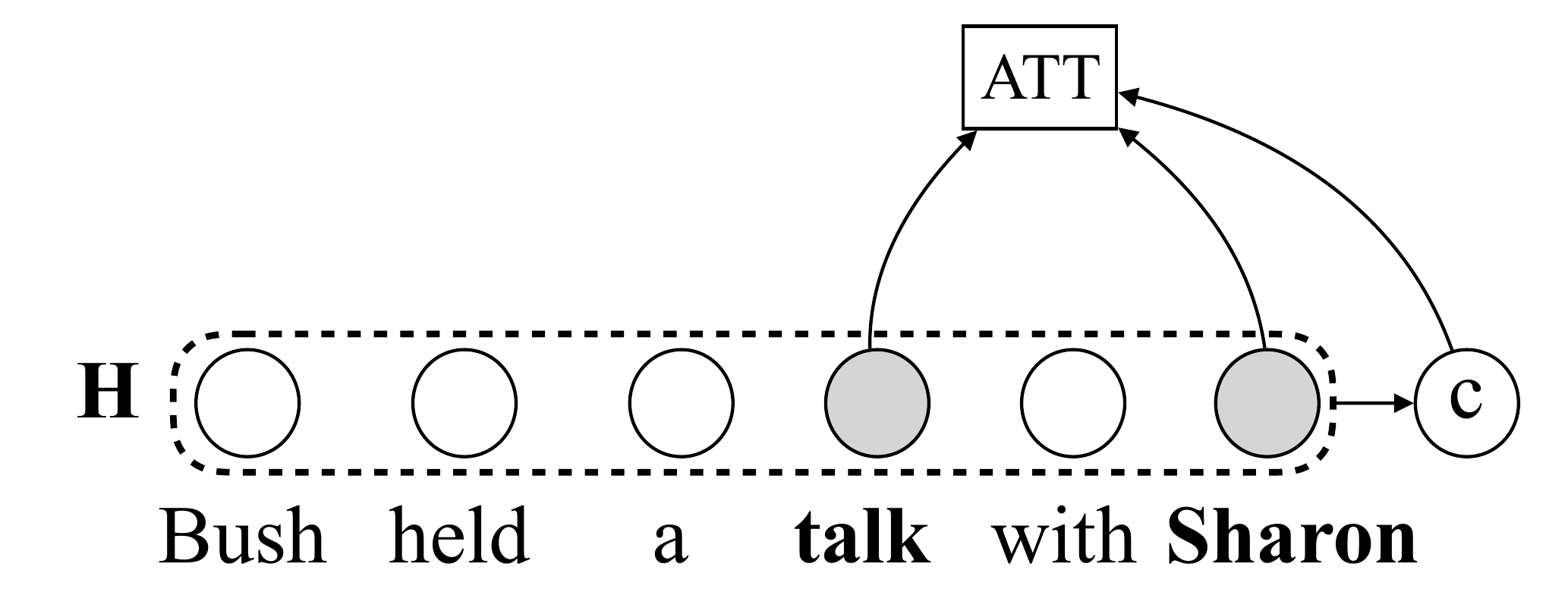}}\vfill
\subfloat[Deep Context]{\includegraphics[width=0.7\columnwidth]{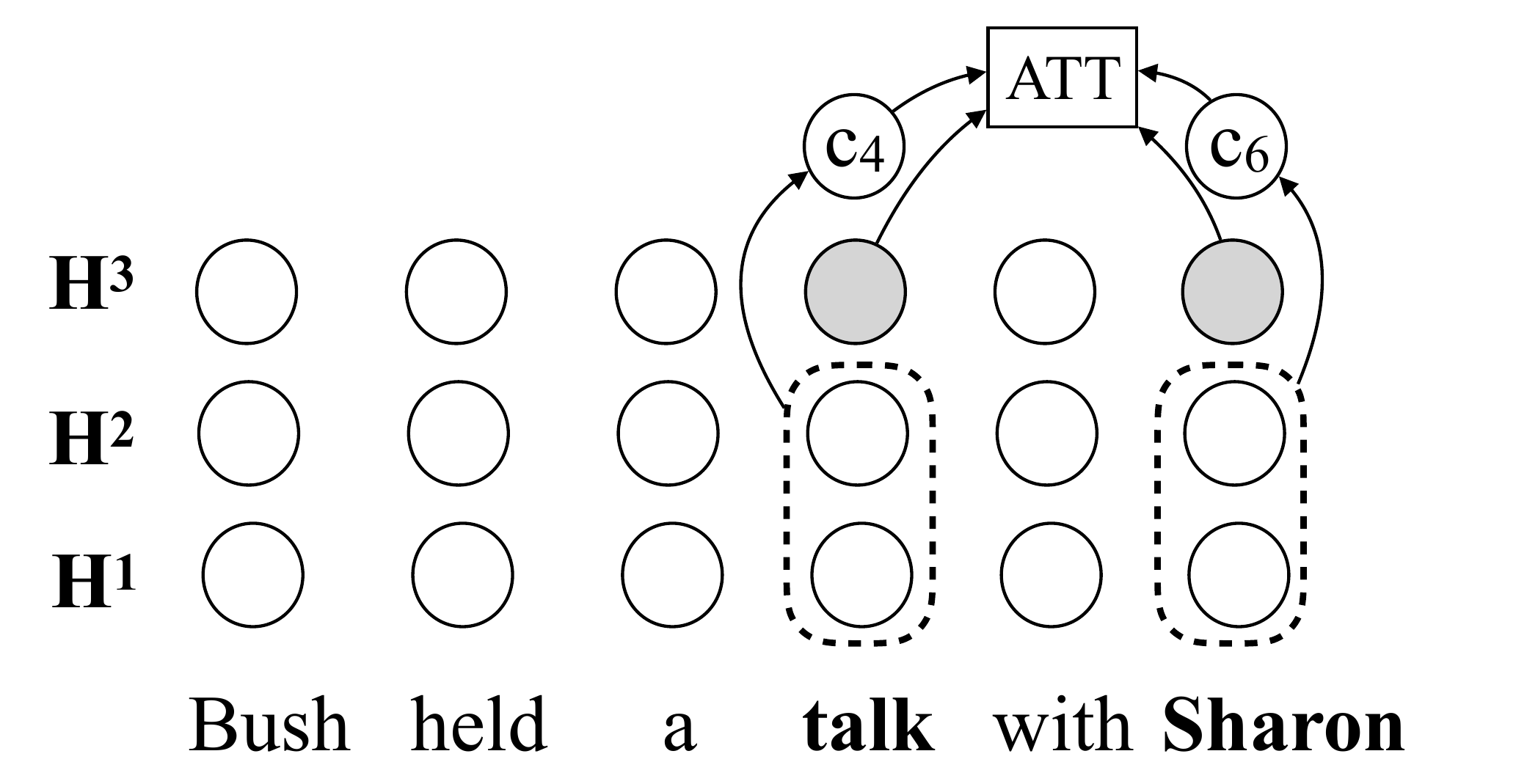}} \hspace{0.2\columnwidth}
\subfloat[Deep-Global Context]{\includegraphics[width=0.7\columnwidth]{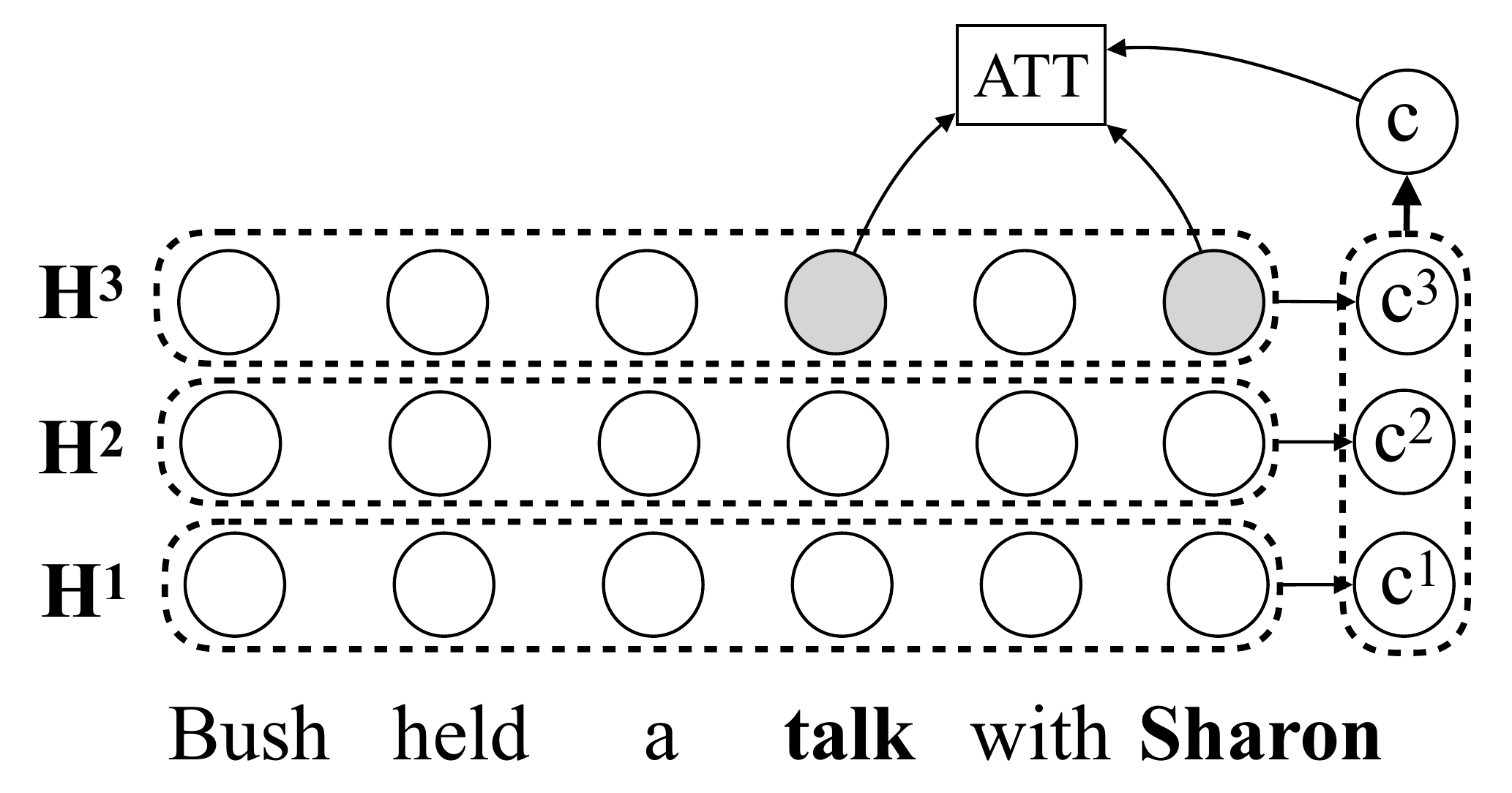}}
\caption{Illustration of the proposed models. As seen, the conventional self-attention networks (a) individually calculate the attention weight of two items (``talk'' and ``sharon'') without covering the contextual information. The global context strategy (b) and the deep-context strategy (c) capture the global meaning of a sentence and the syntactic information from the lower layers respectively. Figure (d) shows a deep-global context model which summarizes the historical global context vectors.}
\label{fig:self-attention}
\end{center}
\end{figure*}

The strength of SANs lies in the ability of directly capturing dependencies between layer hidden states~\cite{Vaswani:2017:NIPS}.
However, the calculation of similarity  between query and key in the self-attention model is merely controlled by two trained parameter matrices:
\begin{eqnarray}
    {\bf Q} {\bf K}^T = ({\bf H} {\bf W}_Q) ({\bf H} {\bf W}_K)^T = {\bf H} ({\bf W}_Q {\bf W}_K^T) {\bf H}^T, \label{eq:energy}
\end{eqnarray}
which miss the opportunity to take advantage of useful contexts. For example, as seen in Figure~\ref{fig:self-attention} (a), self-attention model individually calculate the relevance between the word pair ("talk" and "Sharon") without considering the contextual information. We expect that modeling context can further improve the performance of SANs. 


\section{Approach}
In this study, we propose a context-aware self-attention model. We describe several types of context vectors and introduce how to incorporate the context vectors into the SAN-based sequence-to-sequence models. 

\subsection{Context-Aware Self-Attention Model}

In order to alleviate the lack of contextual information and to maintain the flexibility on parallel computation for self-attention networks, we propose to contextualize the transformations from the input layer $\bf H$ to the query and key layers. Specifically, we follow~\citeauthor{shaw2018self}~\shortcite{shaw2018self} to propagate contextual information to transformation using addition, which avoids significantly increasing computation:


\begin{equation}
  \begin{bmatrix}
   {\bf \widehat{Q}}\\ 
    {\bf \widehat{K}} 
  \end{bmatrix} =  (1-
    \begin{bmatrix}
     {\bf \lambda}_Q\\ 
    {\bf \lambda}_K 
  \end{bmatrix})
  \begin{bmatrix}
    {\bf Q}\\ {\bf K}
  \end{bmatrix} + \begin{bmatrix}
     {\bf \lambda}_Q\\ 
    {\bf \lambda}_K 
  \end{bmatrix}({\bf C}
  \begin{bmatrix}
    {\bf U}_{Q}\\{\bf U}_{K}
  \end{bmatrix}),
\label{eq:context-linear}
\end{equation}
where ${\bf C} \in \mathbb{R}^{n \times d_c}$ is the context vector, and $\{{\bf U}_{Q}, {\bf U}_{K}\} \in \mathbb{R}^{d_c \times d}$ are the associated trainable parameter matrices. To effectively leverage these hierarchical representations,  $\{\lambda_Q, \lambda_K\}\in\mathbb{R}^{n \times 1}$ are assigned to weight the expected importance of the context representations.

\citeauthor{britz2017massive}~\shortcite{britz2017massive} and \citeauthor{Vaswani:2017:NIPS}~\shortcite{Vaswani:2017:NIPS} noted that a large magnitude of $Q$ and $K$ may push the softmax function (Equation~\ref{eq:sim}) into regions where it has extremely small gradients. To counteract this effect, $\{\lambda_Q, \lambda_K\}$ can also be treated as factors to regulate the magnitude of $\widehat{Q}$ and $\widehat{K}$.\footnote{We conduct experiments on the effectiveness of the factors. The experimental results reveal that without $\{\lambda_Q, \lambda_K\}$, there is a big drop (-5.23 BLEU) on the final translation qualities. This indicates that the large magnitude of ${\bf Q}$ and ${\bf K}$ exactly hinder the convergence of SANs, and the trainable linear projections (Equation~\ref{eq:context-linear}) insufficiently learn to regular the magnitude. } Inspired by the prior studies on multi-modal networks \cite{xu2015show,calixto2017doubly,yang2017towards},
we assign a gating scalar to learn the factors:


\begin{equation}
      \begin{bmatrix}
     {\bf \lambda}_Q\\ 
    {\bf \lambda}_K 
  \end{bmatrix} =  \sigma(
  \begin{bmatrix}
    {\bf Q}\\ {\bf K}
  \end{bmatrix}
     \begin{bmatrix}
    {\bf V}_{Q}^{H}\\ {\bf V}_{K}^{H}
  \end{bmatrix}  + 
  {\bf C}
  \begin{bmatrix}
    {\bf U}_{Q}\\{\bf U}_{K}
  \end{bmatrix}
    \begin{bmatrix}
    {\bf V}_Q^C\\ {\bf V}_K^C
  \end{bmatrix}
  ),
\label{eq:context-gate}
\end{equation}
where $\{{\bf V}_Q^H, {\bf V}_K^H\}\in\mathbb{R}^{d \times 1}$ and $\{{\bf V}_Q^C, {\bf V}_K^C\}\in\mathbb{R}^{d_c \times 1}$ are trainable parameters. $\sigma(\cdot)$ denotes the logistic sigmoid function. The gating scalar enables the model to explicitly quantify how much each representation and the context vector contribute to the prediction of attention weight.

Accordingly, the output representation is constructed based on the contextualized query and key representations:
\begin{eqnarray}
   {\bf O} &=& \textsc{Att}({\bf \widehat{Q}}, {\bf \widehat{K}}) \  {\bf V} \label{eq:context-out}.
\end{eqnarray}
As seen, the proposed approach does not require specific attention functions, thus is applicable to all attention models.


\subsection{Choices of Context Vectors}

One principle of our approach is to maintain the simplicity and flexibility of the self-attention model.
With this in mind, we employ the internal states as context vectors, thus avoid relying on external resources. 
Specifically, we exploit several types of context vectors, which can either be used individually or combined together.


\subsubsection{Global Context}
Global context is a function of the entire input layer, which represents the global meaning of a sequence. In this work, we use mean operation to summarize the representations of the input layer, which is commonly used in Seq2Seq models~\cite{cho2014learning}:
\begin{align}
{\bf c} = \overline{\bf H}  && \in \mathbb{R}^d
\label{eqn:global}
\end{align}
Note that the global context is a vector instead of a matrix, which is shared across layer states.

Intuitively, the global context can be regarded as an instance-specific bias \cite{hariharan2015hypercolumns} for the self-attention model, which is expected to complement the unified parameters $\{{\bf W}_Q, {\bf W}_K\}$ shared across instances in the training data.  
The pair-wised features in conjunction with the global features produce an instance-specific prior which has been shown its effectiveness on several recognition and detection tasks \cite{hariharan2015hypercolumns,gkioxari2015contextual,zhu2016beyond}.

\subsubsection{Deep Context}
Deep context is a function of the internal layers stacked below the current input layer.
Advanced neural models are generally implemented as multiple layers, which are able to capture different types of syntactic and semantic information~\cite{shi2016does,Peters:2018:NAACL,Anastasopoulos:2018:NAACL}. For example,~\citeauthor{Peters:2018:NAACL}~\shortcite{Peters:2018:NAACL} show that higher-level layer states capture the context-dependent aspects of word meaning while lower-level states model the aspects of syntax, and simultaneously exposing all of these signals is highly beneficial.

Formally, let ${\bf H}^l$ be the current input layer at the $l$-th level, the deep context is a concatenation of the layers underneath the input layer: 
\begin{align}
{\bf C} = [{\bf H}^1, \dots, {\bf H}^{l-1}]  && \in \mathbb{R}^{n \times (l-1)d}
\label{eqn:deep}
\end{align}
The deep context enables the self-attention model to fuse different types of syntactic and semantic information captured by different layers.

Note that we employ a dense connection strategy~\cite{Huang:2017:CVPR} instead of linear combination~\cite{Peters:2018:NAACL}. We believe the former is a more suitable strategy in this scenario, since the weight matrices $\{{\bf U}_{Q}, {\bf U}_{K}\} \in \mathbb{R}^{(l-1)d \times d}$ in Equation~\ref{eq:context-linear} plays the role of combination. 
Our strategy differs from~\citeauthor{Peters:2018:NAACL}~\shortcite{Peters:2018:NAACL} at: (1) they use normalized weights while we directly use parameters that could be either positive or negative numbers, which may benefit from more modeling flexibility;
(2) they use a scalar that is shared by all elements in the layer states, while we assign a distinct ``scalar'' to each element. The latter offers a more precise control of the combination by allowing the model to be more expressive than scalars~\cite{Tu:2017:TACL}.




\begin{table*}[t]
  \centering
  \begin{tabular}{c|c||c|l||rcc||c}
   {\bf \#}  &   {\bf Model} &  {\bf Applied to}  &  {\bf Context Vectors} &    \bf \# Para.    &   {\bf Train} &  {\bf Decode} &  {\bf  BLEU}\\
    \hline \hline
    1   &   {\textsc{Base}}  &  n/a   &  n/a  & 88.0M   & 1.28 & 1.52 & 27.31\\
    \hline\hline 
    2   &   \multirow{7}{*}{\textsc{Ours}}  &   \multirow{4}{*}{encoder}  &   {\em global context} &  91.0M &   1.26 & 1.50 &  27.96\\ 
    3   &                                   &   &   {\em deep-global context} &  99.0M & 1.25  & 1.48 & 28.15\\ 

   4   &                                   &   &   {\em deep context}  & 95.9M & 1.18 & 1.38 & 28.01 \\   
        5   &                                   &   &   {\em deep-global context} + {\em deep context}  & 106.9M & 1.16 & 1.36 & \textbf{28.26}  \\  
    \cline{3-8}
    6   &                                   &   \multirow{2}{*}{decoder}    &   {\em deep-global context}    & 99.0M & 1.23 & 1.44 &  27.94  \\ 
    7   &                                   &   &   {\em deep-global context} + {\em deep context}   & 106.9M & 1.15 & 1.35 & 28.02   \\   
    \cline{3-8}
    8  &                                   &   both    & 5  + 7    & 125.8M  &  1.04 & 1.20 &  28.16  \\ 
  \end{tabular}
  \caption{Experimental results on WMT14 En$\Rightarrow$De translation task using \textsc{Transformer-Base}. ``\# Para." denotes the trainable parameter size of each model (M = million). ``Train'' and ``Decode' denote the training speed (steps/second) and the decoding speed (sentences/second), respectively.}
  \label{tab:res}
\end{table*}
\subsubsection{Deep-Global Context}
Intuitively, we can combine the concepts of global and deep context, and fuse global context across layers:
\begin{align}
    {\bf c} = [{\bf c}^1, \dots, {\bf c}^{l}]  && \in \mathbb{R}^{ld}
    \label{eqn:deep-global}
\end{align}
where ${\bf c}^l$ is the global context of the $l$-th layer ${\bf H}^l$, which is calculated via Equation~\ref{eqn:global}. We expect the deep-global context to provide different levels of linguistic biases, ranging from lexical, through syntactic, to semantic levels.

As seen, the above context vectors embed different types of information, either global or state-wise context, which may be complementary to each other. To exploit the advantages of all of them, 
an intuitive strategy is to concatenate multiple context vectors to form a new vector, which serves as $\bf C$ in Equation~\ref{eq:context-linear}. 
The proposed approach can be easily integrated into the state-of-the-art SAN-based \textsc{Seq2Seq} models~\cite{Vaswani:2017:NIPS}, in which both encoder and decoder are composed of a stack of $L$ SAN layers.\footnote{For decoder-side SANs, global context vector is a summarization of the forward representations at each decoding step, since the subsequent representations are invisible and thus are masked during training.}

\section{Experiments}
\subsection{Setup}
Following \cite{Vaswani:2017:NIPS}, we evaluates the proposed approach on machine translation tasks. 
To compare with the results reported by previous SAN-based NMT models ~\cite{Vaswani:2017:NIPS,hassan2018achieving}, we conducted experiments on both English$\Rightarrow$German (En$\Rightarrow$De) and Chinese$\Rightarrow$English (Zh$\Rightarrow$En) translation tasks. 
For the En$\Rightarrow$De task, we trained on the widely-used WMT14 dataset consisting of about $4.56$ million sentence pairs.  The models were validated on  newstest2013 and examined on newstest2014. 
For the Zh$\Rightarrow$En task, the models were trained using all of the available parallel corpus from WMT17 dataset, consisting of about $20.62$ million sentence pairs. We used newsdev2017 as the development set and newstest2017 as the test set. 
The English and German sentences were tokenized using the scripts provided in Moses. 
Then, all tokenized sentences were processed by byte-pair encoding (BPE) to alleviate the Out-of-Vocabulary problem \cite{sennrich2015neural} with 32K merge operations for both language pairs.
We used BLEU score \cite{papineni2002bleu} as the evaluation metric.

We evaluated the proposed approaches on our re-implemented \textsc{Transformer} model~\cite{Vaswani:2017:NIPS}. 
We followed~\cite{Vaswani:2017:NIPS} to set the configurations and reproduced their reported results on the En$\Rightarrow$De task.
We tested both the \emph{Base} and \emph{Big} models, which differ at the layer size (512 vs. 1024) and the number of attention heads (8 vs. 16). All the models are trained on eight NVIDIA P40 GPUs, each of which is allocated a batch of 4096 tokens.  In consideration of the computation cost, we studied the variations of the \emph{Base} model on En$\Rightarrow$De task, and evaluated the overall performance with the \emph{Big} model on both En$\Rightarrow$De and Zh$\Rightarrow$En translation tasks.

\begin{table*}[t]
\begin{center}
\begin{tabular}{l|l||rrl|rrl}
    \multirow{2}{*}{\bf System}  &   \multirow{2}{*}{\bf Architecture}  & \multicolumn{3}{c}{\bf Zh$\Rightarrow$En}  &  \multicolumn{3}{|c}{\bf En$\Rightarrow$De}\\
    \cline{3-8}
        &   &   \# Para. & Train  &   BLEU    &   \# Para.  & Train  &   BLEU\\
    \hline \hline
    \multicolumn{8}{c}{{\em Existing NMT systems}} \\
    \hline
    \multirow{2}{*}{\cite{Vaswani:2017:NIPS}} &   \textsc{Transformer-Base}    & n/a & n/a & n/a &  65M &  n/a &   27.30\\ 
    &  \textsc{Transformer-Big}               &  n/a  & n/a &  n/a  &  213M &  n/a  &  28.40\\ 
    \hdashline
    \cite{hassan2018achieving}  &   \textsc{Transformer-Big}  &  n/a &  n/a  &  24.20  &  n/a  &  n/a  & n/a\\
    \hline\hline
    \multicolumn{8}{c}{{\em Our NMT systems}}   \\ \hline
    \multirow{4}{*}{\em this work}  &   \textsc{Transformer-Base}  &    107.9M & 1.21 & 24.13  &  88.0M & 1.28  &  27.31\\ 
    &   ~~~ + Context-Aware SANs                    & 126.8M & 1.10 & 24.67$^\Uparrow$ & 106.9M & 1.16  & 28.26$^\Uparrow$  \\ 
    \cline{2-8}
    &   \textsc{Transformer-Big}	                & 303.9M  & 0.58 &  24.56 &  264.1M & 0.61 & 28.58   \\ 
    &   ~~~ + Context-Aware SANs                     & 379.4M & 0.41 & 25.15$^\Uparrow$    &339.6M & 0.44 & 28.89 \\ 
  \end{tabular}
  \caption{Comparing with the existing NMT systems on WMT17 Zh$\Rightarrow$En and WMT14 En$\Rightarrow$De test sets.  ``$\Uparrow$'': significant over the conventional self-attention counterpart ($p < 0.01$), tested by bootstrap resampling.} 
  \label{tab:exist}
  \end{center}
\end{table*}
\begin{figure*}[t]
\begin{center}
\includegraphics[width=2.1\columnwidth]{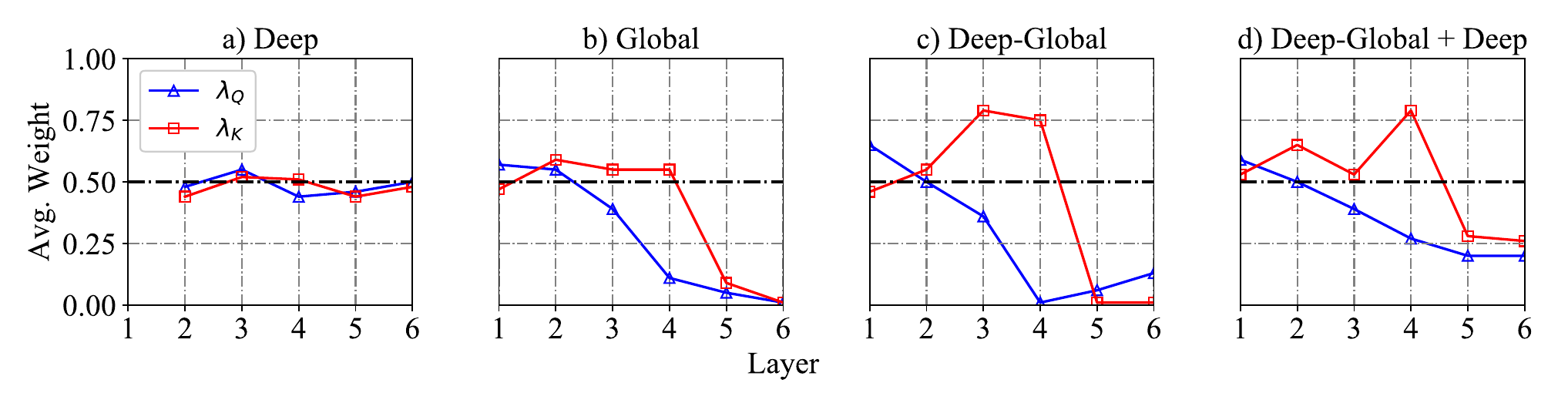}
\caption{Visualization of the importance of each type of context vector on different layers. The importance is assessed by averaging the scalar factors in Equation~\ref{eq:context-linear} over the validation set and distinguished by ${\lambda}_Q$ (blue) and ${\lambda}_K$ (red).}
\label{fig:importance-layer}
\end{center}
\end{figure*}
\subsection{Ablation Study on the Context Vector}

In this section, we conducted experiments to evaluate the impact of different types of context vector on the WMT14 En$\Rightarrow$De translation tasks using the \textsc{Transformer-Base}.
First, we investigated the effect of different context vectors for the encoder-side self-attention networks.
Then, we examined whether modeling contextual information on the decoder-side SANs is able to gain consistent improvement.
Finally, we checked whether the context-aware model on encoder-side and decoder-side SANs can be complementary to each other. To eliminate the influence of control variables, we conducted the first two ablation studies on encoder-side or decoder-side self-attention networks only. 

\subsubsection{Applied to Encoder} As shown in Table~\ref{tab:res}, all the proposed context vector strategies consistently improve the model performance  over  the  baseline, validating  the  importance of modeling contextual information in self-attention networks. Among them,  global context (model \#2)  and  deep context (model \#4) gained comparable improvements. Deep-global context outperforms its global counterpart, showing that the different levels of global linguistic biases benefit to the accuracy of translation. Moreover, we evaluated whether the global and deep manners are complementary to each other. By simply summing them to the final context vectors, the model ``deep-global context + deep context'' (model \#5) gains further improvements. According to the results, we argue that the two types of context vectors are able to improve the SANs in different aspects. 

\subsubsection{Applied to Decoder}
In this group of experiments, we investigated the question of which types of context vector should be applied to the decoder-side self-attention networks. As shown in Table~\ref{tab:res}, both ``deep-global context'' (model \#6) and ``deep-global context + deep context'' (model \#7) consistently improve the SANs. Still, the later outperforms the former one, which is same to the phenomenon appeared in the experiments in terms of encoder. Noted that, ~\citeauthor{zhang2018accelerating}~\shortcite{zhang2018accelerating} pointed out that the decoder-side SANs tends to only focus on its nearby representation. However, our improvements show that all the forward (global) and lower layer (deep) representations are still necessary for the decoder-side SANs.  



\subsubsection{Applied to Both Encoder and Decoder}
Finally, we integrated the strategies into both the encoder and decoder. As seen, this strategy (model \#8)  
even slightly harms the translation quality (compare to encoder-side models). We attribute the drop of BLEU score 
to the fact that the conventional encoder-decoder attention model in \textsc{Transformer} exploits the top-layer of encoder representations, which already embeds useful contextual information. The context-aware model may benefit more on encoder-side SAN under the architecture of \textsc{Transformer}.

Unless otherwise stated, considering the training speed, we only applied the context-aware model to the encoder-side SANs in the following experiments, which employs a ``deep-global context + deep context '' strategy (Row \#5 in Table~\ref{tab:res}).

\subsection{Main Results}
Table~\ref{tab:exist} lists the results on WMT17 Zh$\Rightarrow$En and WMT14 En$\Rightarrow$De translation tasks. Our baseline models, both \textsc{Transformer-Base} and \textsc{Transformer-Big}, outperform the reported results on the same data, which makes the evaluation convincing. As seen, modeling contextual information (``Context-Aware SANs'') consistently improves the performance across language pairs and model variations, demonstrating the efficiency and universality of the proposed approach. It is encouraging to see that \textsc{Transformer-Base} with context-aware model achieves comparable performance with \textsc{Transformer-Big}, while the training speed is nearly twice faster and only requires half of parameters.

\section{Analysis}
We conducted extensive analyses to better understand our model in terms of their compatibility with self-attention networks. All the results are reported on En$\Rightarrow$De validation set with \textsc{Transformer-Base}.
\subsection{Deep Context vs. Global Context}
In this section, we investigated the details of difference between deep context and global context to answer the question: how do they harmonically work with queries and keys in multiple layers? 

\subsubsection{Stable Necessity of Deep Context}
As seen in Figure~\ref{fig:importance-layer} (a), in deep based context-aware models, the weights of scalar factors are consistently close to 0.5, meaning the equivalent importance of the information in the current layer and that of the historical layers.  
The improvements on the translation quality and the stable necessity jointly verified our claim that the conventional self-attention mechanism  is insufficient to fully capture the richness of the context through weighted averaging its input layer. 
Opportunely utilizing the historical context benefits the performance of SANs. 

\subsubsection{The Lower Layer, The More Global Context Required} Concerning the global-based approaches, obviously,  the average weights of global context vectors in Figure~\ref{fig:importance-layer} (b), (c) and (d) drop at high-level layers. The trends demonstrate that the higher layers require less global information, while the lower layers require more global context. The phenomenon confirms that a single SAN layer has limited ability in learning the global meaning, resulting in the high weights of global context vector in lower layers. However the global contextual information can be gradually accumulated with the increasing number of layers, this is the reason why the higher layers hardly need the global information. We believe the global context vector is more beneficial to the lower layers on modeling semantic meanings. 

\begin{table}[h]
  \centering
  \setcounter{table}{2}
  \begin{tabular}{c|c|c||c}
   \bf Model &   \bf Query   &   \bf Key    &  \bf  Dev\\
    \hline \hline
    \textsc{Transformer-Base}  &   -   &  - &  25.84 \\ \hline 
    \multirow{3}{*}{   ~~~ + Context-Aware}  &	\checkmark   & 	\checkmark    & \bf 26.42 \\ 
              &	\texttimes   &  \checkmark   & 26.36 \\ 
             &  \checkmark   &  \texttimes   & 26.20 \\ 
  \end{tabular}
  \caption{BLEU scores on the En$\Rightarrow$De validation set with
respect to integrate context vector into queries and keys. }
  \label{tab:qk}
\end{table}

\subsubsection{Keys Required More Global Information}
Another common interesting phenomenon appears in all the global-based approaches is that the  weights of global context vectors for keys are usually higher than that for queries, especially in the mid-level layers. 
 We believe this is caused by the different usage of query and key. Considering the normalization in $softmax$ function (Equation~\ref{eq:sim}) which is effected on the keys, each key should consider its relationships to other items. This is why the keys require more semantic information in SANs. The results in Table~\ref{tab:qk} show that self-attention networks indeed benefit more from incorporating global information into keys than that of queries. 
 However, it should be noted that enhancing the queries with context representations can further improves the performance. 

\subsection{Source Context vs. Target Context}
\begin{figure}[h]
\begin{center}
\includegraphics[width=0.85\columnwidth]{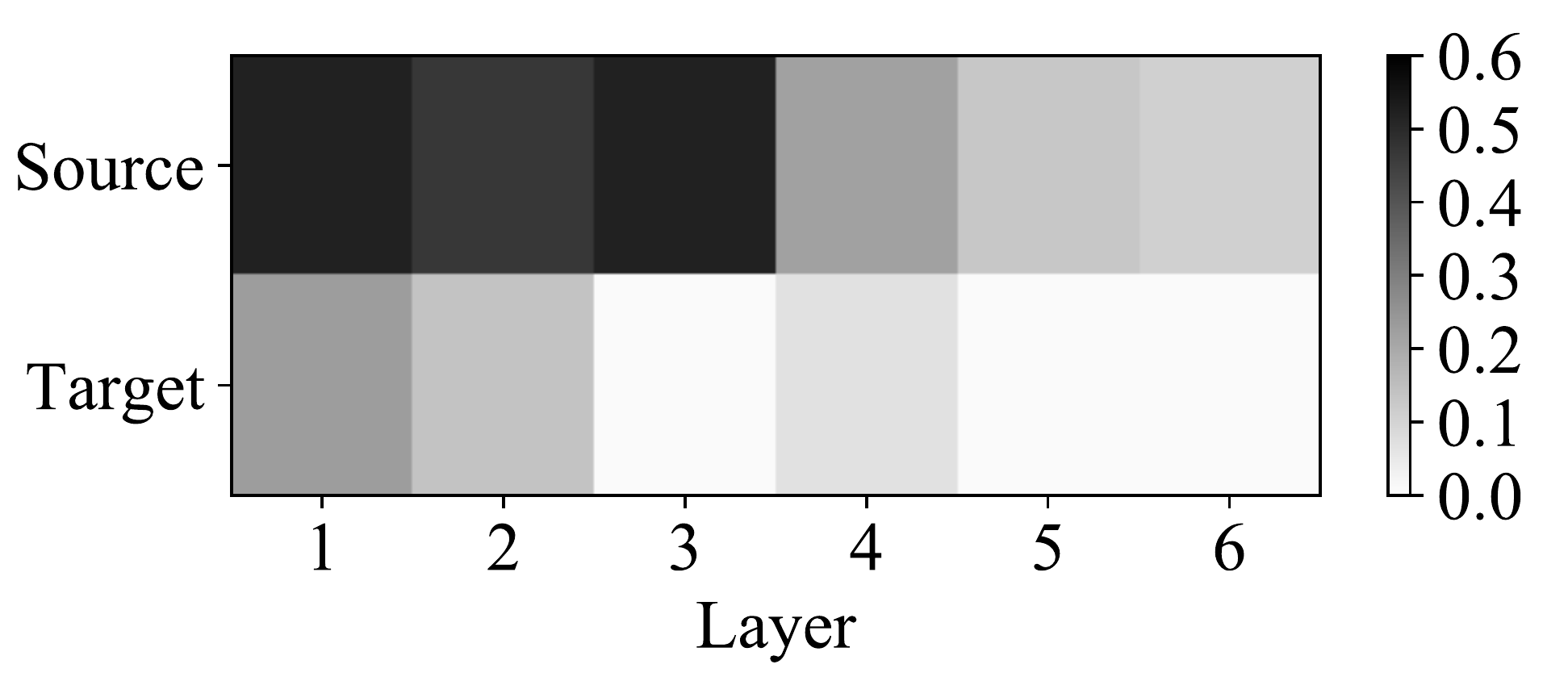}
\caption{Visualization of weights learned for source-side context vectors and target-side context vectors when we integrate the context-aware model on the both sides of \textsc{Transformer}. Obviously, target context vectors are allocated with much lower weights than its source-side counterpart.}
\label{fig:dec-weight}
\end{center}
\end{figure}
In this part, we investigated why integrating encoder-side and decoder-side context vectors fails to  further improve the self-attention model. We took a deep look into the Model \#8 (See Table~\ref{tab:res}), and averaged the gating scalar of the source-side and target-side context  vectors, respectively. As observed in Figure~\ref{fig:dec-weight}, the target-side context vectors consistently gain minor weights, resulting in less contribution from the target-side contextual information. Concerning the source-side context vectors, the factors automatically allocate larger proportion. The result verified our claim that the top-layer of encoder representations has already embedded with useful contextual information, which has exploited to the target-side representations through the conventional encoder-decoder attention network. Thus, the decoder-side context-aware model does not further improve the translation quality as expected.   

\subsection{Linguistic Analysis}
The last, we provided linguistic analyses to the proposed models in terms of: 1) whether the proposed model is flexibly as expected to utilize the contextual information for different words; and 2) how the proposed models perform on sentences of different lengths. 
\subsubsection{Analysis on Part-of-Speech}
\begin{figure}[h]
\begin{center}
\includegraphics[width=0.85\columnwidth]{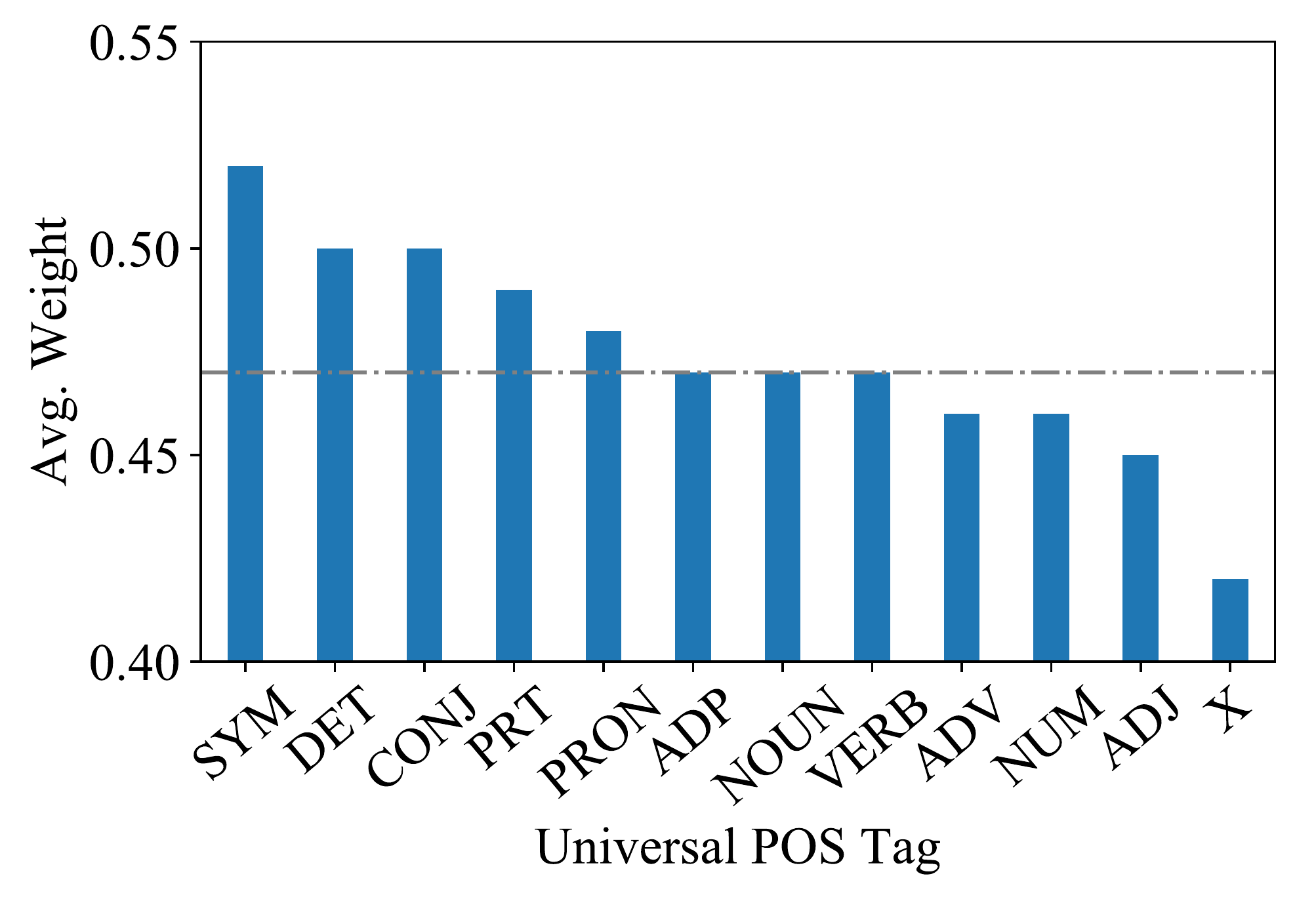}
\caption{The weights of deep-global context vectors corresponding to different POS.  
Grey line indicates the average weight of all the words. As observed, the function words require more contextual information than content words.}
\label{fig:tagging}
\end{center}
\end{figure}
Since the context representations are element-wised added to the SAN model, an interesting question is whether different words are assigned with distinct weights. We categorized the words in validation set using the Part-of-Speech (POS) tag set.\footnote{Including: ``SYM''-symbols, `DET''- determiner, ``CONJ''-conjuntion, ``PRT''-partical, ``PRON''-pronoun, ``ADP''-adposition, ``NOUN''-noun, ``VERB''-verb, ``ADV''-adverb, ``Num''-number, ``ADJ''-adjective, and ``X''-others.}
Figure~\ref{fig:tagging} shows the factors learned for controlling the weight of context vectors. The function words, which have very little substantive meaning (e.g, ``SYM'', `DET'', ``CONJ'', ``PRT'', ``PRON'' and ``ADP''), require more contextual information than that of content words such as the nouns, verbs, adjectives, and adverbs.  
We attribute the phenomena to the fact that the representations of function words profit more from contextual information, which also noted by \citeauthor{wang2018learning}~\shortcite{wang2018learning} who suggested to reconstruct the function words (e.g. the pronouns) to alleviate the problem of dropped pronoun.

\subsubsection{Analysis on Long Sentences}
\begin{figure}[t]
\begin{center}
\includegraphics[width=0.90\columnwidth]{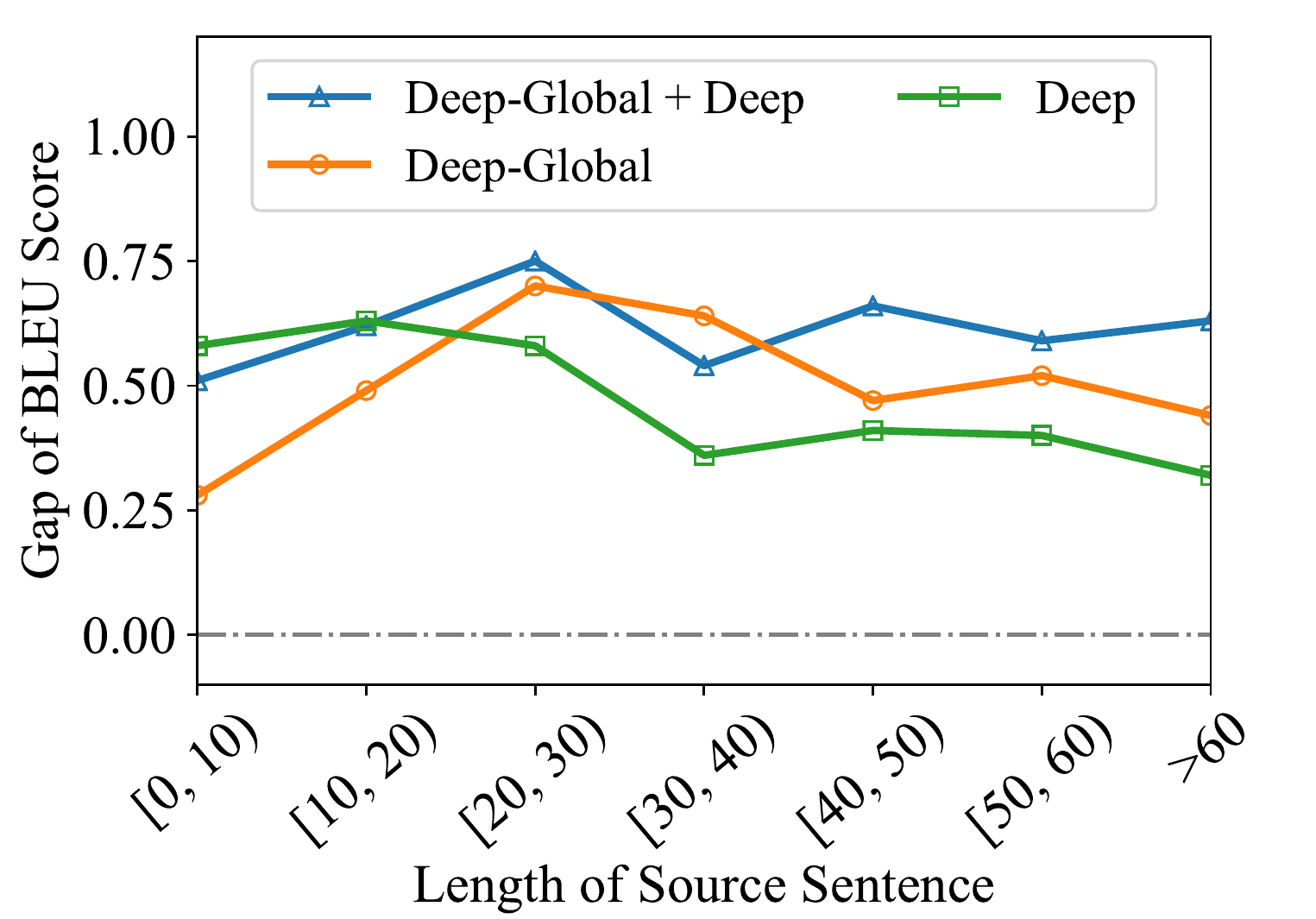}
\caption{Performance improvement according to  various input sentence lengths. Y-axis denotes the gap of BLEU score between our model and baseline (grey line).}
\label{fig:length-bleu}
\end{center}
\end{figure}

We followed~\citeauthor{tu2016modeling}~\shortcite{tu2016modeling} to evaluate the effect of context-aware models on long sentences. 
The sentences were divided into 10 disjoint groups according to their lengths and their BLEU scores were evaluated, as shown in Figure~\ref{fig:length-bleu}. 

The proposed approaches outperform the baseline model in almost all the length segments. There are still considerable differences between the global-based and the deep-based variations. Global-based strategies consistently outperform the deep-context model on sentences with more than 20 words, while the opposite situation appears on the shorter sentences.  One possible reason is that translating long sentences require more long-distance dependency information which can be supplemented by the global contextual information. For short sentences, the effect of global context is relatively minor, while the complex syntactic and semantic dependencies from deep context provide more impact on the translation quality.

\section{Related Work}
Neural representations embed complex characteristics of word use (e.g., syntax and semantic)~\cite{Choi:2016:CSL}. Several researchers have shown that contextual information can enhance the ability of modeling dependencies among neural representations, especially for the attention models~\cite{bahdanau2015neural,luong2015effective}. For example,  \citeauthor{Tu:2017:TACL}~\shortcite{Tu:2017:TACL} and \citeauthor{zhang2017context}~\shortcite{zhang2017context} respectively enhanced the query and memory of a conventional attention model with internal contextual representations. Their studies verified the necessity  of global contextual information for modeling the dependencies between representations. Moreover,~\citeauthor{Peters:2018:NAACL}~\shortcite{Peters:2018:NAACL} pointed out that deeply modeling syntactic and semantic contexts from multiple layers benefits to the performance of multi-layer neural networks.   Contrary to the prior studies explored on RNN-based approaches or required external resources, our work focuses on improving the self-attention networks with contextual information.



Although self-attention model has shown its strength in modeling discrete sequence on different tasks, e.g machine translation~\cite{Vaswani:2017:NIPS}, natural language inference~\cite{Shen:2018:AAAI} and acoustic modeling~\cite{sperber2018self}, several studies have mentioned the limitations in conventional self-attention networks~\cite{Chen:2018:ACL}. 
Among them,~\citeauthor{Yang:2018:EMNLP}~\shortcite{Yang:2018:EMNLP} noted that restricting the attention model to a local space may benefit to the performance, supporting that the conventional SAN model insufficiently fully capturing the context of a sequence. 
\citeauthor{tang2018self}~\shortcite{tang2018self} found that the conventional SANs fail to fully take the advantages of direct connections between elements. Moreover, \citeauthor{shaw2018self}~\shortcite{shaw2018self} succeed on incorporating relative positions into the SAN models, supporting that the conventional model requires necessary information for modeling relations between the states. \citeauthor{bawden2018evaluating}~\shortcite{bawden2018evaluating} and \citeauthor{Voita:2018:ACL}~\shortcite{Voita:2018:ACL} enhanced the attention network  with external contextual representations that summarizes previous source sentences. Unlike their work which requires the embeddings of previous sentences, our approaches contextualize the transformations in SANs, thus avoid relying on external resources and maintain the simplicity and flexibility.      


\section{Conclusion}
In this work, we improved the self-attention networks with contextual information. We proposed several simple but effective strategies to model the contexts, and found that the deep and global approaches are complementary to each other. Experimental results across language pairs demonstrate the effectiveness and universality of the proposed approach. Extensive analyses show that how the context-aware model enhances the original representations in the self-attention model, and our model is able to flexibly model the contextual information for different representations.   

It is interesting to validate the model in other tasks, such as reading comprehension, language inference, and stance classification~\cite{Xu:2018:ACL}.
Another promising direction is to design 
more linguistic context-aware techniques, such as incorporating the linguistic knowledge (e.g. phrases and syntactic categories).
It is also interesting to combine with other techniques~\cite{shaw2018self,Li:2018:EMNLP,Dou:2018:EMNLP,Dou:2019:AAAI,Kong:2019:AAAI,yang2018convolutional} to further boost the performance of Transformer.

\section*{Acknowledgments}
This work was supported in part by the National Natural Science Foundation of China (Grant No. 61672555), the Joint Project of Macao Science and Technology Development Fund and National Natural Science Foundation of China (Grant No. 045/2017/AFJ) and the Multiyear Research Grant from the University of Macau (Grant No. MYRG2017-00087-FST).
We thank the anonymous reviewers for their insightful comments.

\bibliography{main}
\bibliographystyle{aaai}

\end{document}